# N-norm and N-conorm in Neutrosophic Logic and Set, and the Neutrosophic Topologies

Florentin Smarandache
University of New Mexico, Gallup
NM 87301, USA
E-mail: smarand@unm.edu

**Abstract**:
In this paper we present the N-norms/N-conorms in neutrosophic logic and set as extensions of T-norms/T-conorms in fuzzy logic and set.
Also, as an extension of the Intuitionistic Fuzzy Topology we present the Neutrosophic Topologies.

1. **Definition of the Neutrosophic Logic/Set**:

Let T, I, F be real standard or non-standard subsets of $]^{-}0, 1^{+}[$,
with sup T = t_sup, inf T = t_inf,
sup I = i_sup, inf I = i_inf,
sup F = f_sup, inf F = f_inf,
and n_sup = t_sup+i_sup+f_sup,
n_inf = t_inf+i_inf+f_inf.

Let U be a universe of discourse, and M a set included in U. An element x from U is noted with respect to the set M as x(T, I, F) and belongs to M in the following way: it is t% true in the set, i% indeterminate (unknown if it is or not) in the set, and f% false, where t varies in T, i varies in I, f varies in F.

Statically T, I, F are subsets, but dynamically T, I, F are functions/operators depending on many known or unknown parameters.

2. In a similar way define the **Neutrosophic Logic**:
A logic in which each proposition x is T% true, I% indeterminate, and F% false, and we write it x(T,I,F), where T, I, F are defined above.

3. As a generalization of T-norm and T-conorm from the Fuzzy Logic and Set, we now introduce the **N-norms and N-conorms** for the Neutrosophic Logic and Set.

We define a *partial relation order* on the neutrosophic set/logic in the following way:
$x(T_1, I_1, F_1) \leq y(T_2, I_2, F_2)$ iff (if and only if) $T_1 \leq T_2$, $I_1 \geq I_2$, $F_1 \geq F_2$ for crisp components.

And, in general, for subunitary set components:
$x(T_1, I_1, F_1) \leq y(T_2, I_2, F_2)$ iff

$\inf T_1 \leq \inf T_2$, $\sup T_1 \leq \sup T_2$,
$\inf I_1 \geq \inf I_2$, $\sup I_1 \geq \sup I_2$,
$\inf F_1 \geq \inf F_2$, $\sup F_1 \geq \sup F_2$.

If we have mixed - crisp and subunitary - components, or only crisp components, we can transform any crisp component, say "a" with a $\in [0,1]$ or a$\in ]^-0, 1^+[$, into a subunitary set [a, a]. So, the definitions for subunitary set components should work in any case.

### 3.1. N-norms

$N_n: ( ]^-0,1^+[ \times ]^-0,1^+[ \times ]^-0,1^+[ )^2 \rightarrow ]^-0,1^+[ \times ]^-0,1^+[ \times ]^-0,1^+[$

$N_n (x(T_1,I_1,F_1), y(T_2,I_2,F_2)) = (N_nT(x,y), N_nI(x,y), N_nF(x,y))$,
where $N_nT(.,.), N_nI(.,.), N_nF(.,.)$ are the truth/membership, indeterminacy, and respectively falsehood/nonmembership components.

$N_n$ have to satisfy, for any x, y, z in the neutrosophic logic/set M of the universe of discourse U, the following axioms:
a) Boundary Conditions: $N_n(x, \mathbf{0}) = \mathbf{0}$, $N_n(x, \mathbf{1}) = x$.
b) Commutativity: $N_n(x, y) = N_n(y, x)$.
c) Monotonicity: If $x \leq y$, then $N_n(x, z) \leq N_n(y, z)$.
d) Associativity: $N_n(N_n (x, y), z) = N_n(x, N_n(y, z))$.

There are cases when not all these axioms are satisfied, for example the associativity when dealing with the neutrosophic normalization after each neutrosophic operation. But, since we work with approximations, we can call these **N-pseudo-norms**, which still give good results in practice.

$N_n$ represent the *and* operator in neutrosophic logic, and respectively the *intersection* operator in neutrosophic set theory.

Let $J \in \{T, I, F\}$ be a component.
Most known N-norms, as in fuzzy logic and set the T-norms, are:
• The Algebraic Product N-norm: $N_{n-algebraic}J(x, y) = x \cdot y$
• The Bounded N-Norm: $N_{n-bounded}J(x, y) = \max\{0, x + y - 1\}$
• The Default (min) N-norm: $N_{n-min}J(x, y) = \min\{x, y\}$.

A general example of N-norm would be this.
Let $x(T_1, I_1, F_1)$ and $y(T_2, I_2, F_2)$ be in the neutrosophic set/logic M. Then:
$$N_n(x, y) = (T_1 \wedge T_2, I_1 \vee I_2, F_1 \vee F_2)$$
where the "$\wedge$" operator, acting on two (standard or non-standard) subunitary sets, is a N-norm (verifying the above N-norms axioms); while the "$\vee$" operator, also acting on two (standard or non-standard) subunitary sets, is a N-conorm (verifying the below N-conorms axioms).

For example, $\wedge$ can be the Algebraic Product T-norm/N-norm, so $T_1 \wedge T_2 = T_1 \cdot T_2$ (herein we have a product of two subunitary sets – using simplified notation); and $\vee$ can be the Algebraic Product T-conorm/N-conorm, so $T_1 \vee T_2 = T_1 + T_2 - T_1 \cdot T_2$ (herein we have a sum, then a product, and afterwards a subtraction of two subunitary sets).

Or $\wedge$ can be any T-norm/N-norm, and $\vee$ any T-conorm/N-conorm from the above and below; for example the easiest way would be to consider the *min* for crisp components (or *inf* for subset components) and respectively *max* for crisp components (or *sup* for subset components).

If we have crisp numbers, we can at the end neutrosophically normalize.

### 3.2. N-conorms

$$N_c: (\,]^-0,1^+[\, \times\, ]^-0,1^+[\, \times\, ]^-0,1^+[\,)^2 \to\, ]^-0,1^+[\, \times\, ]^-0,1^+[\, \times\, ]^-0,1^+[$$

$$N_c\,(x(T_1,I_1,F_1),\, y(T_2,I_2,F_2)) = (N_cT(x,y),\, N_cI(x,y),\, N_cF(x,y)),$$

where $N_nT(.,.)$, $N_nI(.,.)$, $N_nF(.,.)$ are the truth/membership, indeterminacy, and respectively falsehood/nonmembership components.

$N_c$ have to satisfy, for any x, y, z in the neutrosophic logic/set M of universe of discourse U, the following axioms:
a) Boundary Conditions: $N_c(x, \mathbf{1}) = \mathbf{1}$, $N_c(x, \mathbf{0}) = x$.
b) Commutativity: $N_c\,(x, y) = N_c(y, x)$.
c) Monotonicity: if $x \leq y$, then $N_c(x, z) \leq N_c(y, z)$.
d) Associativity: $N_c\,(N_c(x, y), z) = N_c(x, N_c(y, z))$.

There are cases when not all these axioms are satisfied, for example the associativity when dealing with the neutrosophic normalization after each neutrosophic operation. But, since we work with approximations, we can call these **N-pseudo-conorms**, which still give good results in practice.

$N_c$ represent the *or* operator in neutrosophic logic, and respectively the *union* operator in neutrosophic set theory.

Let $J \in \{T, I, F\}$ be a component.
Most known N-conorms, as in fuzzy logic and set the T-conorms, are:
• The Algebraic Product N-conorm: $N_{c-algebraic}J(x, y) = x + y - x \cdot y$
• The Bounded N-conorm: $N_{c-bounded}J(x, y) = \min\{1, x + y\}$
• The Default (max) N-conorm: $N_{c-max}J(x, y) = \max\{x, y\}$.

A general example of N-conorm would be this.
Let $x(T_1, I_1, F_1)$ and $y(T_2, I_2, F_2)$ be in the neutrosophic set/logic M. Then:
$$\mathbf{N_n(x, y) = (T_1 \vee T_2,\, I_1 \wedge I_2,\, F_1 \wedge F_2)}$$
Where – as above - the "$\wedge$" operator, acting on two (standard or non-standard) subunitary sets, is a N-norm (verifying the above N-norms axioms); while the "$\vee$" operator, also acting on two (standard or non-standard) subunitary sets, is a N-conorm (verifying the above N-conorms axioms).

For example, $\wedge$ can be the Algebraic Product T-norm/N-norm, so $T_1 \wedge T_2 = T_1 \cdot T_2$ (herein we have a product of two subunitary sets); and $\vee$ can be the Algebraic Product T-conorm/N-conorm, so $T_1 \vee T_2 = T_1 + T_2 - T_1 \cdot T_2$ (herein we have a sum, then a product, and afterwards a subtraction of two subunitary sets).

Or $\wedge$ can be any T-norm/N-norm, and $\vee$ any T-conorm/N-conorm from the above; for example the easiest way would be to consider the *min* for crisp components (or *inf* for subset components) and respectively *max* for crisp components (or *sup* for subset components).

If we have crisp numbers, we can at the end neutrosophically normalize.

Since the min/max (or inf/sup) operators work the best for subunitary set components, let's present their definitions below. They are extensions from subunitary intervals {defined in [3]} to any subunitary sets. Analogously we can do for all neutrosophic operators defined in [3].

Let $x(T_1, I_1, F_1)$ and $y(T_2, I_2, F_2)$ be in the neutrosophic set/logic M.

**Neutrosophic Conjunction/Intersection**:
$x \wedge y = (T_\wedge, I_\wedge, F_\wedge)$,

where $\inf T_\wedge = \min\{\inf T_1, \inf T_2\}$
$\sup T_\wedge = \min\{\sup T_1, \sup T_2\}$
$\inf I_\wedge = \max\{\inf I_1, \inf I_2\}$
$\sup I_\wedge = \max\{\sup I_1, \sup I_2\}$
$\inf F_\wedge = \max\{\inf F_1, \inf F_2\}$
$\sup F_\wedge = \max\{\sup F_1, \sup F_2\}$

**Neutrosophic Disjunction/Union**:
$x \vee y = (T_\vee, I_\vee, F_\vee)$,
where $\inf T_\vee = \max\{\inf T_1, \inf T_2\}$
$\sup T_\vee = \max\{\sup T_1, \sup T_2\}$
$\inf I_\vee = \min\{\inf I_1, \inf I_2\}$
$\sup I_\vee = \min\{\sup I_1, \sup I_2\}$
$\inf F_\vee = \min\{\inf F_1, \inf F_2\}$
$\sup F_\vee = \min\{\sup F_1, \sup F_2\}$

**Neutrosophic Negation/Complement:**
$\mathcal{C}(x) = (T_\mathcal{C}, I_\mathcal{C}, F_\mathcal{C})$,
where $T_\mathcal{C} = F_1$
$\inf I_\mathcal{C} = 1 - \sup I_1$
$\sup I_\mathcal{C} = 1 - \inf I_1$
$F_\mathcal{C} = T_1$

Upon the above Neutrosophic Conjunction/Intersection, we can define the

**Neutrosophic Containment:**
We say that the neutrosophic set A is included in the neutrosophic set B of the universe of discourse U,
iff for any $x(T_A, I_A, F_A) \in A$ with $x(T_B, I_B, F_B) \in B$ we have:
$\inf T_A \leq \inf T_B$ ; $\sup T_A \leq \sup T_B$;
$\inf I_A \geq \inf I_B$ ; $\sup I_A \geq \sup I_B$;
$\inf F_A \geq \inf F_B$ ; $\sup F_A \geq \sup F_B$.

3.3. **Remarks**.
a) The non-standard unit interval $]^-0, 1^+[$ is merely used for philosophical applications, especially when we want to make a distinction between relative truth (truth in at least one world) and absolute truth (truth in all possible worlds), and similarly for distinction between relative or absolute falsehood, and between relative or absolute indeterminacy.

But, for technical applications of neutrosophic logic and set, the domain of definition and range of the N-norm and N-conorm can be restrained to the normal standard real unit interval [0, 1], which is easier to use, therefore:

$N_n: ([0,1] \times [0,1] \times [0,1])^2 \rightarrow [0,1] \times [0,1] \times [0,1]$

and

$N_c: ([0,1] \times [0,1] \times [0,1])^2 \rightarrow [0,1] \times [0,1] \times [0,1]$.

b) Since in NL and NS the sum of the components (in the case when T, I, F are crisp numbers, not sets) is not necessary equal to 1 (so the normalization is not required), we can keep the final result un-normalized.

But, if the normalization is needed for special applications, we can normalize at the end by dividing each component by the sum all components.

If we work with intuitionistic logic/set (when the information is incomplete, i.e. the sum of the crisp components is less than 1, i.e. *sub-normalized*), or with paraconsistent logic/set (when the information overlaps and it is contradictory, i.e. the sum of crisp components is greater than 1, i.e. *over-normalized*), we need to define the neutrosophic measure of a proposition/set.

If x(T,I,F) is a NL/NS, and T,I,F are crisp numbers in [0,1], then the **neutrosophic vector norm** of variable/set x is the sum of its components:

$$N_{vector-norm}(x) = T+I+F.$$

Now, if we apply the $N_n$ and $N_c$ to two propositions/sets which maybe intuitionistic or paraconsistent or normalized (i.e. the sum of components less than 1, bigger than 1, or equal to 1), x and y, what should be the neutrosophic measure of the results $N_n(x,y)$ and $N_c(x,y)$ ?

Herein again we have more possibilities:
- either the product of neutrosophic measures of x and y:
  $N_{vector-norm}(N_n(x,y)) = N_{vector-norm}(x) \cdot N_{vector-norm}(y)$,
- or their average:
  $N_{vector-norm}(N_n(x,y)) = (N_{vector-norm}(x) + N_{vector-norm}(y))/2$,
- or other function of the initial neutrosophic measures:

  $N_{vector-norm}(N_n(x,y)) = f(N_{vector-norm}(x), N_{vector-norm}(y))$, where f(.,.) is a function to be determined according to each application.

Similarly for $N_{vector-norm}(N_c(x,y))$.

Depending on the adopted neutrosophic vector norm, after applying each neutrosophic operator the result is neutrosophically normalized. We'd like to mention that "**neutrosophically normalizing**" doesn't mean that the sum of the resulting crisp components should be 1 as in fuzzy logic/set or intuitionistic fuzzy logic/set, but the sum of the components should be as above: either equal to the product of neutrosophic vector norms of the initial propositions/sets, or equal to the neutrosophic average of the initial propositions/sets vector norms, etc.

In conclusion, we neutrosophically normalize the resulting crisp components T`,I`,F` by multiplying each neutrosophic component T`,I`,F` with S/( T`+I`+F`), where
S= $N_{vector-norm}(N_n(x,y))$ for a N-norm or S= $N_{vector-norm}(N_c(x,y))$ for a N-conorm - as defined above.

c) If T, I, F are subsets of [0, 1] the problem of neutrosophic normalization is more difficult.
   i) If sup(T)+sup(I)+sup(F) < 1, we have an *intuitionistic proposition/set*.
   ii) If inf(T)+inf(I)+inf(F) > 1, we have a *paraconsistent proposition/set*.
   iii) If there exist the crisp numbers t ∈ T, i ∈ I, and f ∈ F such that t+i+f =1, then we can say that we have a *plausible normalized proposition/set*.
   But in many such cases, besides the normalized particular case showed herein, we also have crisp numbers, say $t_1 \in T$, $i_1 \in I$, and $f_1 \in F$ such that $t_1+i_1+f_1 < 1$ (incomplete information) and $t_2 \in T$, $i_2 \in I$, and $f_2 \in F$ such that $t_2+i_2+f_2 > 1$ (paraconsistent information).

## 4. Examples of Neutrosophic Operators which are N-norms or N-pseudonorms or, respectively N-conorms or N-pseudoconorms.

We define a binary **neutrosophic conjunction (intersection)** operator, which is a particular case of a N-norm (neutrosophic norm, a generalization of the fuzzy T-norm):

$$c_N^{TIF} : ([0,1] \times [0,1] \times [0,1])^2 \to [0,1] \times [0,1] \times [0,1]$$
$$c_N^{TIF}(x,y) = (T_1T_2, I_1I_2 + I_1T_2 + T_1I_2, F_1F_2 + F_1I_2 + F_1T_2 + F_2T_1 + F_2I_1).$$

The neutrosophic conjunction (intersection) operator $x \wedge_N y$ component truth, indeterminacy, and falsehood values result from the multiplication
$$(T_1 + I_1 + F_1) \cdot (T_2 + I_2 + F_2)$$
since we consider in a prudent way $T \prec I \prec F$, where "$\prec$" is a **neutrosophic relationship** and means "weaker", i.e. the products $T_iI_j$ will go to $I$, $T_iF_j$ will go to $F$, and $I_iF_j$ will go to $F$ for all i, j ∈ {1,2}, i≠j, while of course the product $T_1T_2$ will go to T, $I_1I_2$ will go to I, and $F_1F_2$ will go to F (or reciprocally we can say that $F$ prevails in front of $I$ which prevails in front of $T$, and this neutrosophic relationship is transitive):

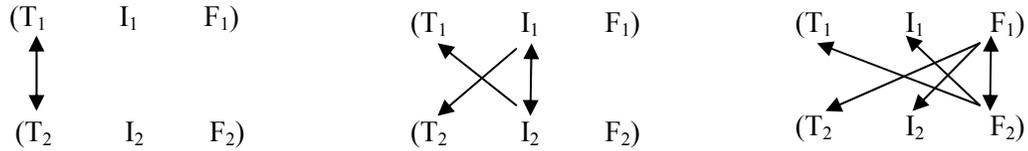

So, the truth value is $T_1T_2$, the indeterminacy value is $I_1I_2 + I_1T_2 + T_1I_2$ and the false value is $F_1F_2 + F_1I_2 + F_1T_2 + F_2T_1 + F_2I_1$. The norm of $x \wedge_N y$ is $(T_1 + I_1 + F_1) \cdot (T_2 + I_2 + F_2)$. Thus, if $x$ and $y$ are normalized, then $x \wedge_N y$ is also normalized. Of course, the reader can redefine the neutrosophic conjunction operator, depending on application, in a different way, for example in a more optimistic way, i.e. $I \prec T \prec F$ or $T$ prevails with respect to $I$, then we get:
$$c_N^{ITF}(x,y) = (T_1T_2 + T_1I_2 + T_2I_1, I_1I_2, F_1F_2 + F_1I_2 + F_1T_2 + F_2T_1 + F_2I_1).$$
Or, the reader can consider the order $T \prec F \prec I$, etc.

Let's also define the unary **neutrosophic negation** operator:
$$n_N : [0,1] \times [0,1] \times [0,1] \to [0,1] \times [0,1] \times [0,1]$$
$$n_N(T, I, F) = (F, I, T)$$
by interchanging the truth $T$ and falsehood $F$ vector components.

Similarly, we now define a binary **neutrosophic disjunction (or union)** operator, where we consider the neutrosophic relationship $F \prec I \prec T$:
$$d_N^{FIT} : ([0,1] \times [0,1] \times [0,1])^2 \to [0,1] \times [0,1] \times [0,1]$$
$$d_N^{FIT}(x,y) = (T_1T_2 + T_1I_2 + T_2I_1 + T_1F_2 + T_2F_1, I_1F_2 + I_2F_1 + I_1I_2, F_1F_2)$$

We consider as neutrosophic norm of the neutrosophic variable $x$, where $NL(x) = T_1 + I_1 + F_1$, the sum of its components: $T_1 + I_1 + F_1$, which in many cases is 1, but can also be positive <1 or >1.

Or, the reader can consider the order $F \prec T \prec I$, in a pessimistic way, i.e. focusing on indeterminacy I which prevails in front of the truth T, or other **neutrosophic order** of the neutrosophic components T,I,F depending on the application.
Therefore,
$$d_N^{FTI}(x,y) = (T_1T_2 + T_1F_2 + T_2F_1, I_1F_2 + I_2F_1 + I_1I_2 + T_1I_2 + T_2I_1, F_1F_2)$$

### 4.1. Neutrophic Composition k-Law

Now, we define a more general neutrosophic composition law, named k-law, in order to be able to define neutrosophic k-conjunction/intersection and neutrosophic k-disjunction/union for k variables, where $k \geq 2$ is an integer.

Let's consider $k \geq 2$ neutrosophic variables, $x_i(T_i, I_i, F_i)$, for all $i \in \{1, 2, ..., k\}$. Let's denote
$$T = (T_1, ..., T_k)$$
$$I = (I_1, ..., I_k)$$
$$F = (F_1, ..., F_k)$$

We now define a neutrosophic composition law $o_N$ in the following way:
$$o_N : \{T, I, F\} \to [0,1]$$

If $z \in \{T, I, F\}$ then $z o_N z = \prod_{i=1}^{k} z_i$.

If $z, w \in \{T, I, F\}$ then
$$z o_N w = w o_N z = \sum_{\substack{r=1 \\ \{i_1,...,i_r, j_{r+1},...,j_k\} \equiv \{1,2,...,k\} \\ (i_1,...,i_r) \in C^r(1,2,...,k) \\ (j_{r+1},...,j_k) \in C^{k-r}(1,2,...,k)}}^{k-1} z_{i_1} ... z_{i_r} w_{j_{r+1}} ... w_{j_k}$$

where $C^r(1,2,...,k)$ means the set of combinations of the elements $\{1,2,...,k\}$ taken by $r$. [Similarly for $C^{k-r}(1,2,...,k)$.]

In other words, $z o_N w$ is the sum of all possible products of the components of vectors $z$ and $w$, such that each product has at least a $z_i$ factor and at least a $w_j$ factor, and each product has exactly $k$ factors where each factor is a different vector component of $z$ or of $w$. Similarly if we multiply three vectors:
$$T o_N I o_N F = \sum_{\substack{u,v,k-u-v=1 \\ \{i_1,...,i_u, j_{u+1},...,j_{u+v}, l_{u+v+1},...,l_k\} \equiv \{1,2,...,k\} \\ (i_1,...,i_u) \in C^u(1,2,...,k), (j_{u+1},...,j_{u+v}) \in \\ \in C^v(1,2,...,k), (l_{u+v+1},...,l_k) \in C^{k-u-v}(1,2,...,k)}}^{k-2} T_{i_1}...T_{i_u} I_{j_{u+1}}...I_{j_{u+v}} F_{l_{u+v+1}}...F_{l_k}$$

Let's see an example for $k = 3$.

$$x_1(T_1, I_1, F_1)$$
$$x_2(T_2, I_2, F_2)$$
$$x_3(T_3, I_3, F_3)$$
$$T_{o_N}T = T_1T_2T_3, \quad I_{o_N}I = I_1I_2I_3, \quad F_{o_N}F = F_1F_2F_3$$
$$T_{o_N}I = T_1I_2I_3 + I_1T_2I_3 + I_1I_2T_3 + T_1T_2I_3 + T_1I_2T_3 + I_1T_2T_3$$
$$T_{o_N}F = T_1F_2F_3 + F_1T_2F_3 + F_1F_2T_3 + T_1T_2F_3 + T_1F_2T_3 + F_1T_2T_3$$
$$I_{o_N}F = I_1F_2F_3 + F_1I_2F_3 + F_1F_2I_3 + I_1I_2F_3 + I_1F_2I_3 + F_1I_2I_3$$
$$T_{o_N}I_{o_N}F = T_1I_2F_3 + T_1F_2I_3 + I_1T_2F_3 + I_1F_2T_3 + F_1I_2T_3 + F_1T_2I_3$$

For the case when indeterminacy $I$ is not decomposed in subcomponents {as for example $I = P \cup U$ where $P$ =paradox (true and false simultaneously) and $U$ =uncertainty (true or false, not sure which one)}, the previous formulas can be easily written using only three components as:

$$T_{o_N}I_{o_N}F = \sum_{i,j,r \in \mathcal{P}(1,2,3)} T_i I_j F_r$$

where $\mathcal{P}(1,2,3)$ means the set of permutations of $(1,2,3)$ i.e.
$$\{(1,2,3),(1,3,2),(2,1,3),(2,3,1,),(3,1,2),(3,2,1)\}$$

$$z_{o_N}w = \sum_{\substack{i=1 \\ (i,j,r) \equiv (1,2,3) \\ (j,r) \in \mathcal{P}^2(1,2,3)}}^{3} z_i w_j w_{j_r} + w_i z_j z_r$$

This neurotrophic law is associative and commutative.

## 4.2. Neutrophic Logic and Set k-Operators

Let's consider the neutrophic logic crispy values of variables $x, y, z$ (so, for k = 3):
$$NL(x) = (T_1, I_1, F_1) \text{ with } 0 \leq T_1, I_1, F_1 \leq 1$$
$$NL(y) = (T_2, I_2, F_2) \text{ with } 0 \leq T_2, I_2, F_2 \leq 1$$
$$NL(z) = (T_3, I_3, F_3) \text{ with } 0 \leq T_3, I_3, F_3 \leq 1$$

In neutrosophic logic it is not necessary to have the sum of components equals to 1, as in intuitionist fuzzy logic, i.e. $T_k + I_k + F_k$ is not necessary 1, for $1 \leq k \leq 3$

As a particular case, we define the tri-nary conjunction neutrosophic operator:
$$c_{3N}^{TIF} : ([0,1] \times [0,1] \times [0,1])^3 \to [0,1] \times [0,1] \times [0,1]$$
$$c_{3N}^{TIF}(x, y, z) = \left( T_{o_N}T, I_{o_N}I + I_{o_N}T, F_{o_N}F + F_{o_N}I + F_{o_N}T \right)$$

If all x, y, z are normalized, then $c_{3N}^{TIF}(x, y, z)$ is also normalized.

If x, y, or y are non-normalized, then $\left|c_{3N}^{TIF}(x, y, z)\right| = |x| \cdot |y| \cdot |z|$, where |w| means norm of w.

$c_{3N}^{TIF}$ is a 3-N-norm (neutrosophic norm, i.e. generalization of the fuzzy T-norm).

Again, as a particular case, we define the unary negation neutrosophic operator:
$$n_N : [0,1] \times [0,1] \times [0,1] \to [0,1] \times [0,1] \times [0,1]$$
$$n_N(x) = n_N(T_1, I_1, F_1) = (F_1, I_1, T_1).$$

Let's consider the vectors:
$$T = \begin{pmatrix} T_1 \\ T_2 \\ T_3 \end{pmatrix}, \quad I = \begin{pmatrix} I_1 \\ I_2 \\ I_3 \end{pmatrix} \text{ and } F = \begin{pmatrix} F_1 \\ F_2 \\ F_3 \end{pmatrix}.$$

We note $T_{\bar{x}} = \begin{pmatrix} F_1 \\ T_2 \\ T_3 \end{pmatrix}$, $T_{\bar{y}} = \begin{pmatrix} T_1 \\ F_2 \\ T_3 \end{pmatrix}$, $T_{\bar{z}} = \begin{pmatrix} T_1 \\ T_2 \\ F_3 \end{pmatrix}$, $T_{\overline{xy}} = \begin{pmatrix} F_1 \\ F_2 \\ T_3 \end{pmatrix}$, etc.

and similarly
$$F_{\bar{x}} = \begin{pmatrix} T_1 \\ F_2 \\ F_3 \end{pmatrix}, \quad F_{\bar{y}} = \begin{pmatrix} F_1 \\ T_2 \\ F_3 \end{pmatrix}, \quad F_{\overline{xz}} = \begin{pmatrix} T_1 \\ F_2 \\ T_3 \end{pmatrix}, \text{ etc.}$$

For shorter and easier notations let's denote $z_{o_N} w = zw$ and respectively $z_{o_N} w_{o_N} v = zwv$ for the vector neutrosophic law defined previously.

Then the neutrosophic tri-nary conjunction/intersection of neutrosophic variables x, y, and z is:
$$c_{3N}^{TIF}(x, y, z) = (TT, II + IT, FF + FI + FT + FIT) =$$
$$= (T_1 T_2 T_3, I_1 I_2 I_3 + I_1 I_2 T_3 + I_1 T_2 I_3 + T_1 I_2 I_3 + I_1 T_2 T_3 + T_1 I_2 T_3 + T_1 T_2 I_3,$$
$$F_1 F_2 F_3 + F_1 F_2 I_3 + F_1 I_2 F_3 + I_1 F_2 F_3 + F_1 I_2 I_3 + I_1 F_2 I_3 + I_1 I_2 F_3 +$$
$$+ F_1 F_2 T_3 + F_1 T_2 F_3 + T_1 F_2 F_3 + F_1 T_2 T_3 + T_1 F_2 T_3 + T_1 T_2 F_3 +$$
$$+ T_1 I_2 F_3 + T_1 F_2 I_3 + I_1 F_2 T_3 + I_1 T_2 F_3 + F_1 I_2 T_3 + F_1 T_2 I_3).$$

Similarly, the neutrosophic tri-nary disjunction/union of neutrosophic variables x, y, and z is:
$$d_{3N}^{FIT}(x, y, z) = (TT + TI + TF + TIF, II + IF, FF) =$$
$(T_1T_2T_3 + T_1T_2I_3 + T_1I_2T_3 + I_1T_2T_3 + T_1I_2I_3 + I_1T_2I_3 + I_1I_2T_3 + T_1T_2F_3 + T_1F_2T_3 + F_1T_2T_3 + T_1F_2F_3 + F_1T_2F_3 + F_1F_2T_3 + T_1I_2F_3 + T_1F_2I_3 + I_1T_2F_3 + I_1F_2T_3 + F_1I_2T_3 + F_1T_2I_3, I_1I_2I_3 + I_1I_2F_3 + I_1F_2I_3 + F_1I_2I_3 + I_1F_2F_3 + F_1I_2F_3 + F_1F_2I_3, F_1F_2F_3)$

Surely, other neutrosophic orders can be used for tri-nary conjunctions/intersections and respectively for tri-nary disjunctions/unions among the componenets T, I, F.

**5. Neutrosophic Topologies.**

A) General Definition of NT:

Let M be a non-empty set.

Let $x(T_A, I_A, F_A) \in A$ with $x(T_B, I_B, F_B) \in B$ be in the neutrosophic set/logic M, where A and B are subsets of M. Then (see Section 2.9.1 about N-norms / N-conorms and examples):

$A \cup B = \{x \in M, x(T_A \vee T_B, I_A \wedge I_B, F_A \wedge F_B)\}$,
$A \cap B = \{x \in M, x(T_A \wedge T_B, I_A \vee I_B, F_A \vee F_B)\}$,
$\mathcal{C}(A) = \{x \in M, x(F_A, I_A, T_A)\}$.

A General Neutrosophic Topology on the non-empty set M is a family $\eta$ of Neutrosophic Sets in M satisfying the following axioms:

- **0**(0,0,1) and **1**(1,0,0) $\in \eta$;
- If $A, B \in \eta$, then $A \cap B \in \eta$;
- If the family $\{A_k, k \in K\} \subset \eta$, then $\bigcup_{k \in K} A_k \in \eta$.

B) An alternative version of NT

-We cal also construct a Neutrosophic Topology on NT = $]^{-}0, 1^{+}[$, considering the associated family of standard or non-standard subsets included in NT, and the empty set ∅, called open sets, which is closed under set union and finite intersection.

Let A, B be two such subsets. The union is defined as:

$A \cup B = A+B-A \cdot B$, and the intersection as: $A \cap B = A \cdot B$. The complement of A, $C(A) = \{1^{+}\}-A$, which is a closed set. {When a non-standard number occurs at an extremity of an internal, one can write "]" instead of "(" and "[" instead of ")".} The interval NT, endowed with this topology, forms a *neutrosophic topological space*.

In this example we have used the Algebraic Product N-norm/N-conorm. But other Neutrosophic Topologies can be defined by using various N-norm/N-conorm operators.

In the above defined topologies, if all x's are paraconsistent or respectively intuitionistic, then one has a Neutrosophic Paraconsistent Topology, respectively Neutrosophic Intuitionistic Topology.